\documentclass[letterpaper, 10 pt, conference]{ieeeconf}  
\usepackage{amsmath,amsfonts}
\usepackage{algorithm}
\usepackage{algorithmicx}
\usepackage{algpseudocode}
\usepackage{array}
\usepackage[caption=false,font=normalsize,labelfont=sf,textfont=sf]{subfig}
\usepackage{textcomp}
\usepackage{stfloats}
\usepackage{url}
\usepackage{verbatim}
\usepackage{graphicx}
\usepackage{cite}
\usepackage{xcolor}
\usepackage{cleveref}

\IEEEoverridecommandlockouts                              

\title{\LARGE \bf
Online Safety Filter for Deformable Object Manipulation\\ with Horizon Agnostic Neural Operators
}

\author{
Jiaxing Li$^{*1}$, Hanjiang Hu$^{*1}$, Zhuoyuan Wang$^{1}$, 
Yorie Nakahira$^{1}$, Changliu Liu$^{1}$%
\thanks{*Equal contribution.}%
\thanks{$^{1}$All authors are with Carnegie Mellon University, Pittsburgh, PA, USA.
{\tt\small \{jiaxingl, hanjianghu, zhuoyuaw, ynakahir, cliu6\}@andrew.cmu.edu}}%
}
\begin{document}

\maketitle
\thispagestyle{empty}
\pagestyle{empty}






\begin{abstract}
Safety-critical control of robotic manipulation tasks involving deformable media---such as fluids, cloth, and soft objects---remains challenging because existing learning-based approaches encode safety indirectly through reward shaping, which provides no guarantee of constraint satisfaction at deployment. We present a constraint-driven online safety filter for deformable-object manipulation that enforces explicit task-level safety constraints in real time by minimally modifying any nominal control policy. Our approach combines two key components: a horizon-agnostic neural operator that learns the boundary input-output mapping of the underlying PDE dynamics and generalizes across variable rollout lengths without retraining, and a boundary control barrier function that certifies safety at the task-relevant output level via a lightweight quadratic program. The resulting safety constraint is affine in the boundary-input rate, enabling real-time online filtering. We evaluate the proposed method on fluid manipulation tasks in FluidLab, where the filter improves safe-trajectory rates by up to 22\% over unfiltered base policies while also reducing the number of steps required to reach the safe set, demonstrating that constraint-driven safety enforcement is both more reliable and more efficient than reward-shaping approaches.
\end{abstract}



\section{Introduction}

Robots are increasingly deployed in manipulation tasks involving highly deformable media such as fluids, cloth, and soft objects. In these settings, safety is not merely about stability but about satisfying task-level constraints---avoiding spill, maintaining contact conditions, or keeping objects within designated regions---throughout execution. Most existing learning-based approaches for deformable manipulation are reward-driven~\cite{longhini2025unfolding,zhu2022challenges,gu2023survey,lin2021softgym}: policies are trained to maximize task performance with safety encoded indirectly through penalties or data filtering. However, reward shaping does not guarantee constraint satisfaction, and policies often exhibit unsafe transient behaviors under disturbances or distribution shifts. This limitation becomes particularly pronounced when the underlying dynamics are governed by complex distributed phenomena, such as fluid--rigid interaction~\cite{xian2023fluidlab}, where small control deviations can induce large state changes. These challenges call for a \emph{constraint-driven} safety mechanism that operates online, enforces explicit task-level safety sets, and remains compatible with learned policies.

Achieving such constraint-driven safety for deformable manipulation requires two ingredients: a dynamics model that can predict task-relevant outcomes in real time, and a safety filter that can translate those predictions into hard constraints on the robot's control actions. For manipulation tasks involving fluids or other deformable media, the underlying dynamics are governed by partial differential equations (PDEs), where the robot's action enters as a boundary input and safety-relevant signals (e.g., object position, separation distance) are low-dimensional outputs of the temporal-spatial plant~\cite{bhan2024pde,krstic2008boundary}. Neural operators~\cite{lu2021learning,kovachki2023neural,li2020fourier} can act as function-space approximators that learn input-to-output mappings of PDE systems, offering a promising route to fast, generalizable surrogate models for such dynamics~\cite{hu2024safe}. In parallel, control barrier functions (CBFs) provide a principled, modular approach to safety filtering: they minimally modify a nominal control signal to enforce forward invariance of a user-defined safe set, typically via a lightweight quadratic program (QP)~\cite{ames2016control,ames2019control}. However, combining neural-operator dynamics models with CBF-based safety filters for real-time, online use in robotic manipulation remains an open challenge.

In this paper, we address this challenge by developing an \emph{online} constraint-driven safety filter for deformable-object manipulation, built around a horizon-agnostic neural operator and a boundary control barrier function (BCBF). The key insight is that, by training the neural operator on variable-length input prefixes, a \emph{single} model can support safety filtering across arbitrary rollout lengths without retraining, enabling fully online deployment. The BCBF then operates at the boundary-output level, providing real-time safety certification that is compatible with any nominal policy while imposing only a minimal modification to the control signal. Our contributions are as follows:
\begin{itemize}
  \item We develop an online QP safety filter that pairs a time-dependent BCBF with a differentiable neural-operator surrogate. The resulting safety constraint is affine in the boundary input rate, making it solvable in real time.
  \item We introduce a horizon-agnostic training scheme for neural operators, enabling the same model to support online filtering at any rollout length without retraining.
  \item We evaluate the proposed method on complex fluid manipulation tasks, demonstrating that the online constraint-driven filter consistently outperforms both unfiltered base policies and an offline filter baseline by up to 22\% while reducing the number of steps needed to reach the safe set.
\end{itemize}

\section{Related Work}

\subsection{Deformable Object Manipulation in Robotics}
Robotic manipulation of deformable objects---including cloth, fluids, and soft bodies---has attracted significant attention due to its relevance in household assistance, manufacturing, and laboratory automation. Compared to rigid-body manipulation, deformable systems exhibit high-dimensional, nonlinear, and often partially observed dynamics, making modeling and control substantially more challenging~\cite{zhu2022challenges,gu2023survey}. Textile manipulation has been extensively studied, emphasizing variability in material properties, perception under occlusion, and long-horizon planning~\cite{longhini2025unfolding,zhu2022challenges}. Recent benchmarks such as SoftGym~\cite{lin2021softgym} and FluidLab~\cite{xian2023fluidlab} enable reproducible evaluation of learning-based policies for cloth folding, fluid transport, and object gathering tasks. Complementary toolkits for soft and continuum robots, including SoMoGym~\cite{graule2022somogym} and SofaGym~\cite{schegg2023sofagym}, further broaden the spectrum of deformable systems and control interfaces.

Most existing approaches rely on imitation learning or reinforcement learning to optimize task performance through reward maximization~\cite{longhini2025unfolding,zhu2022challenges,gu2023survey,lin2021softgym}. While these methods have demonstrated promising results, safety is typically encoded indirectly via reward shaping, penalty terms, or data filtering. As a result, policies may exhibit unsafe transient behaviors, especially under long-horizon execution, disturbances, or distribution shift. This limitation is particularly acute in fluid manipulation, where small control deviations can induce large, irreversible state changes~\cite{xian2023fluidlab}. Explicit enforcement of task-level safety constraints during online execution remains underexplored in deformable manipulation settings.

\textbf{Research Gap:} Prior work in deformable manipulation focuses primarily on reward-driven policy optimization rather than constraint-driven safety enforcement, motivating the need for online safety mechanisms compatible with learned controllers.

\subsection{Neural Operators for Modeling PDE-Governed Dynamics}
Many deformable manipulation tasks are naturally governed by distributed dynamics that can be modeled by partial differential equations (PDEs), such as fluid--rigid interaction and continuum deformation~\cite{krstic2008boundary,bhan2024pde}. Neural operators, including DeepONet~\cite{lu2021learning} and Fourier neural operator (FNO)~\cite{li2020fourier}, have emerged as powerful tools for learning mappings between function spaces and approximating PDE solution operators~\cite{kovachki2023neural}. These models generalize across discretizations and domain configurations, making them attractive for high-dimensional physical systems. More advanced neural operator architectures include graph-based neural operators~\cite{li2020multipole}, spectral neural operators~\cite{fanaskov2023spectral}, and physics-informed variants that combine data fitting with PDE constraints to achieve parametric PDE learning~\cite{li2024physics}.

Recent work has explored neural operators for accelerating simulation, model-based control synthesis, and data-driven PDE prediction, including bypassing computationally heavy gain calculations in PDE backstepping~\cite{bhan2023neural,krstic2024neural}. However, most applications focus on forward modeling or policy training rather than embedding neural operator models within real-time safety enforcement frameworks. In particular, their integration with constraint-based filtering mechanisms for robotics tasks remains limited.

\textbf{Research Gap:}
While neural operators provide scalable surrogates for distributed dynamics, their use in online safety filtering for robotics manipulation tasks has not been systematically studied.

\subsection{Model-Based Safety Filters for Robotics}
Model-based safety filters provide a principled mechanism for enforcing state or output constraints by minimally modifying nominal control inputs at runtime. Control barrier functions (CBFs)~\cite{ames2016control,ames2019control} have been widely applied to robotic systems to guarantee forward invariance of safe sets, typically by solving a quadratic program at each control step to project nominal actions onto a constraint-satisfying region. Extensions of CBF address high relative-degree constraints~\cite{nguyen2016exponential,xiao2021high}, sampled-data and delay effects~\cite{singletary2020control}, as well as robust and adaptive settings~\cite{garg2024advances}. Reachability-based filters and safety-constrained model predictive control offer complementary perspectives and are surveyed alongside CBF-based methods in~\cite{wabersich2023data,hsu2023safety}. Recent work further explores constructing robust safety filters online in challenging settings~\cite{knoedler2025safety}.

Most existing safety filters are developed for finite-dimensional dynamical systems and assume explicit state-space models. 
Extensions to infinite-dimensional or partially observed systems are more limited, and few works address safety enforcement when dynamics are governed by distributed PDE models \cite{krstic2008boundary,bhan2024pde}. Recent safe PDE boundary control work proposes neural boundary control barrier functions (BCBFs) combined with differentiable neural operators to derive QP-based safety filtering for boundary outputs~\cite{hu2024safe}, providing the theoretical foundation our method builds upon. However, this prior work operates offline with fixed rollout lengths, limiting its applicability to robotics settings that require online, horizon-agnostic deployment.

\textbf{Research Gap:} Existing safety filters largely assume finite-dimensional models and do not address online constraint enforcement for PDE-governed deformable manipulation systems.

\section{Methodology}
\begin{figure}[t]
    \centering
    \includegraphics[width=\columnwidth]{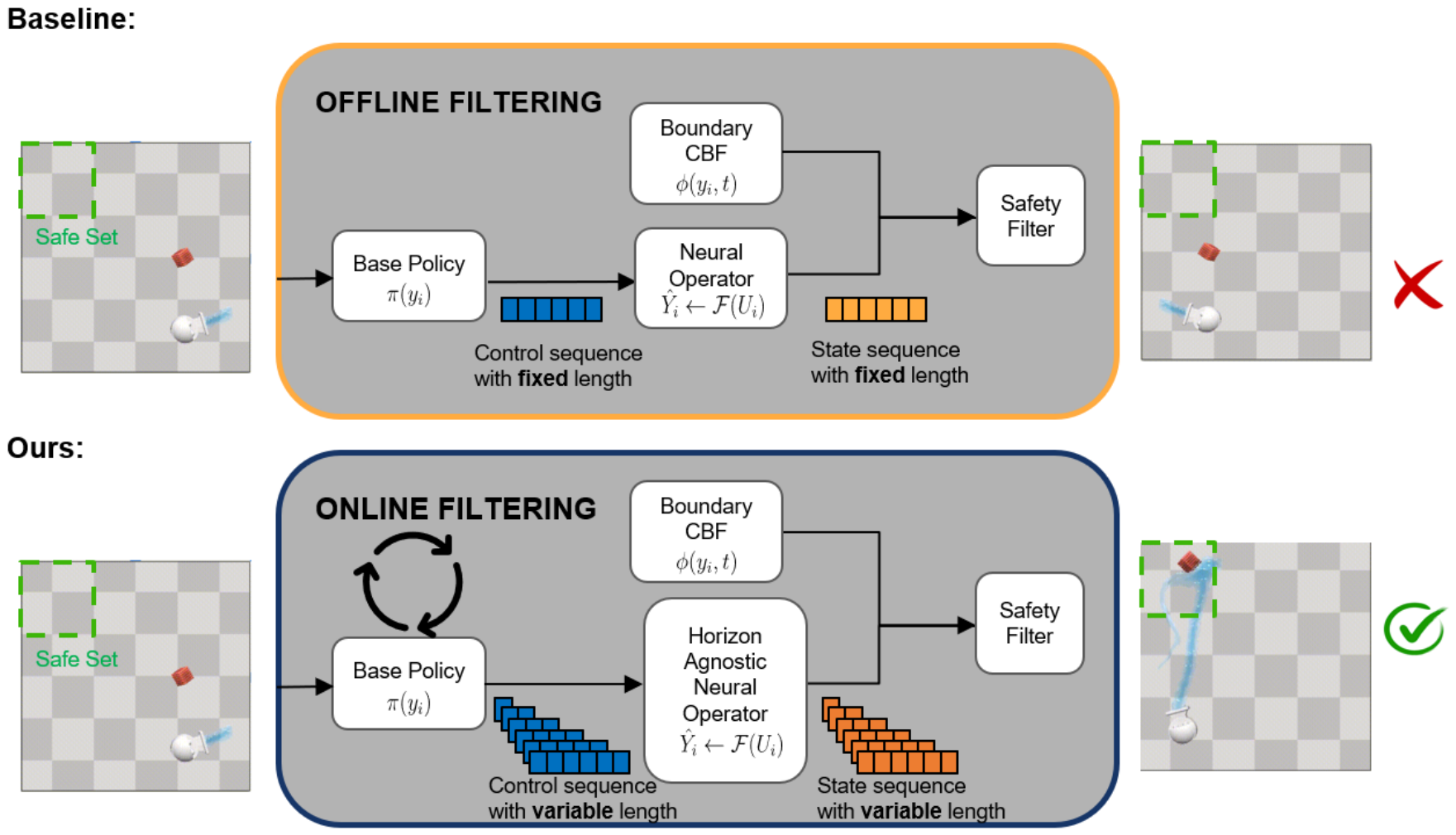}
    \caption{Overall framework of the proposed boundary-safe neural operator control architecture comparing with the offline filtering baseline.}
    \label{fig:flowchar}
\end{figure}
%
\subsection{Problem Formulation}


We consider a general safety-critical control problem for systems governed by unknown
partial differential equations (PDEs), where control is applied through boundary
actuation and safety is specified on task-relevant outputs over a finite horizon.
As a representative instance of this class, we focus on robotic manipulation with
deformable media (e.g., fluid--rigid interaction). In this setting, we denote the
overall coupled state as
$z(x,t): \mathcal{X} \times \mathcal{T} \rightarrow \mathcal{S}$,
where $x \in \mathcal{X}$ is the spatial coordinate, $t \in \mathcal{T} := [0,T]$ is time, and $\mathcal{S}$ is the state space of the coupled robot--environment system.

Similar modeling paradigm is widely adopted in robotics for deformable object manipulation and soft robotic systems, where continuum theories (e.g., elasticity, Cosserat rod models, or fluid dynamics)~\cite{spencer2004continuum} provide PDE descriptions of deformation and interaction with robots.
Following the settings in~\cite{bhan2024pde}, we model the deformable dynamics as
\begin{equation}
\partial_t z(x,t) = \mathcal{D}\!\left(z(x,t), \nabla z(x,t), \nabla^2 z(x,t)\right),
\label{eq:pde_dynamics}
\end{equation}
where $\mathcal{D}(\cdot)$ denotes unknown closed-loop PDE dynamics. The system is actuated through a Dirichlet boundary condition,
\begin{equation}
z(0,t) = U(t),
\label{eq:dirichlet_bc}
\end{equation}
with a fixed initial condition $z(x,0) = z_0(x)$, where $U(t)$ is the control input of the robot agent that is applied to the dynamics of the PDE.

\begin{figure}[h]
    \centering
    \includegraphics[width=0.2\textwidth]{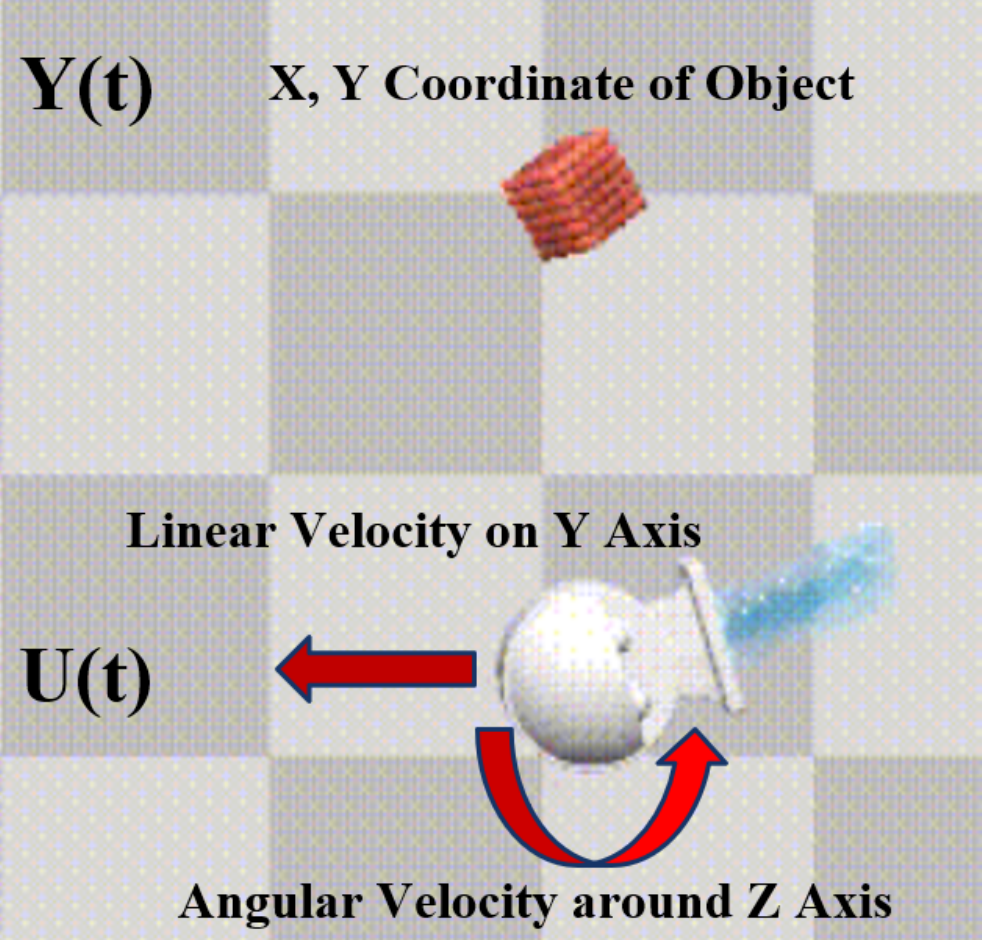}
    \caption{The boundary control input $U$ (linear and angular velocities), and boundary state $Y$ (object's XY coordinate) for the \emph{transporting} task. }
    \label{fig:transport_setting}
\end{figure}

We define the task-relevant boundary output as
\begin{equation}
Y(t) := z(x_b,t),
\end{equation}
where $x_b \in \mathcal{X}$ denotes a designated boundary or observation location of interest. The output $Y(t)$ may represent, as \Cref{fig:transport_setting} shows,
the position of a transported object on XY plane, or a deformation magnitude, or a distance metric derived from the deformable state.



Classical state-dependent control barrier functions (CBFs) are difficult to apply directly in our setting. In PDE-governed robotic manipulation, safety is imposed on a low-dimensional task output \(Y(t)\), while the underlying state \(z(x,t)\) is infinite-dimensional and only partially observed.

To bridge this mismatch, we adopt the boundary control barrier function (BCBF) framework from prior work~\cite{hu2024safe}, which certifies safety directly at the output boundary level. Let \( \mathcal{S}_0 \subset \mathcal{S} \) denote the user-defined safe set for the task output. Boundary-output safety over \( \mathcal{T}=[0,T] \) is characterized by a finite-time feasibility condition: there exists \( t_0\in\mathcal{T} \) such that
\begin{equation}
\forall\, t \in [t_0, T], \quad Y(t) \in \mathcal{S}_0.
\end{equation}
This formulation serves as the theoretical safety foundation of our method. However, existing usage remains offline because direct neural operator-based BCBF construction typically assumes access to full trajectories, which is incompatible with real-time robotics execution.

\subsection{Horizon Agnostic Neural Operator}
We model the boundary input-output mapping using a neural operator
\(
\mathcal{F}_\theta : U(t) \rightarrow Y(t)
\) parameterized by $\theta$,
which learns the transfer function from the boundary control input trajectory
\( U(t) \) to the boundary output trajectory \( Y(t) \).
Following the construction in~\cite{kovachki2023neural}, the neural operator used is defined as
\begin{equation}
\mathcal{F}_\theta
= \mathcal{Q} \circ \mathcal{I}_{L-1} \circ \cdots \circ \mathcal{I}_0 \circ \mathcal{P},
\end{equation}
where \( \mathcal{P} \) and \( \mathcal{Q} \) are pointwise lifting and projection
operators (e.g., Fourier and inverse Fourier transforms in FNO~\cite{li2020fourier}), and each \( \mathcal{I}_\ell \) is an integral kernel operator.
Specifically, the \( \ell \)-th kernel integration layer is defined as
\begin{equation}
\begin{aligned}
v_{\ell+1}(t)
&= \mathcal{I}_\ell(v_\ell)(t) \\
&= \sigma_{\ell+1} \Big(
W_\ell v_\ell(t)
+ \int_{\mathcal{T}}
\kappa^{(\ell)}(t,s)\, v_\ell(s)\, ds
+ b_\ell(t)
\Big),
\end{aligned}
\end{equation}
for \( \ell = 0,\dots,L-1 \), where
\( v_\ell : \mathcal{T} \rightarrow \mathbb{R}^{d_\ell} \) denotes the hidden
representation at layer \( \ell \),
\( \sigma_{\ell+1}(\cdot) \) is a nonlinear activation function,
\( W_\ell \in \mathbb{R}^{d_{\ell+1}\times d_\ell} \) is a local linear operator,
\( \kappa^{(\ell)} \in C(\mathcal{T}\times\mathcal{T};\mathbb{R}^{d_{\ell+1}\times d_\ell}) \)
is a learnable kernel function, and
\( b_\ell(t) \) is a bias term.

To enable real-time safety filtering in robotics without committing to a fixed rollout length, we train a single horizon-agnostic neural operator that is valid over variable-horizon prefixes. This design allows deployment under receding-horizon execution and avoids retraining when task duration changes.
Let
\(\mathcal{D}=\{(U_k,Y_k)\}_{k=1}^{K}\) be trajectory pairs sampled from the unknown
closed-loop PDE dynamics and discretized at \(\{t_m\}_{m=1}^{M}\), where
\(U_k=\{U_k(t_m)\}_{m=1}^{M}\) and \(Y_k=\{Y_k(t_m)\}_{m=1}^{M}\). During training,
we sample a prefix length \(\tilde{M}\in[M_{\min},M_{\max}]\) and feed
\(U_k^{1:\tilde{M}}\) into \(\mathcal{F}_{\theta}\), which induces the horizon-agnostic
empirical risk.
\begin{equation}
\begin{aligned}
\mathcal{L}
&=
\mathbb{E}_{\tilde{M}\sim \mathcal{U}[M_{\min},M_{\max}]}
\Bigg[
\frac{1}{K}\sum_{k=1}^{K}
\frac{1}{\tilde{M}}\sum_{m=1}^{\tilde{M}}  \\
&\qquad
\left\|
Y_k(t_m)
-
\mathcal{F}_{\theta}\!\left(U_k^{1:\tilde{M}}\right)(t_m)
\right\|_2^2
\Bigg].
\end{aligned}
\label{eq:new_la_no_loss}
\end{equation}
Hence, the same \(\mathcal{F}_{\theta}\) is optimized over a distribution of prefix lengths and queried online with ascending-length control histories. In practice, this turns horizon-agnostic learning into a deployment capability for prefix-based real-time safety certification without fixed-horizon assumptions.

\subsection{Online Filtering}

We implement online filtering in a receding one-step manner at each discrete time
\(t_i=i\Delta t\). Let \(u_i\) denote the nominal boundary control from the base policy,
and let \(\hat{u}_i\) denote the filtered control applied to the system.
At step \(i\), the filter keeps the previously applied filtered sequence and appends the current nominal
control \(u_i\), and forms the prefix input
\begin{equation}
\tilde{U}_{0:i} := \{\hat{u}_0,\ldots,\hat{u}_{i-1},u_i\}.
\end{equation}
The neural operator then predicts the boundary-output prefix
\(
\hat{Y}_{0:i}=\mathcal{F}_\theta(\tilde{U}_{0:i})
\),

Using the current observation \(y_i\), we evaluate the BCBF terms
\(\phi_i:=\phi(y_i,t_i)\) and solve the following one-step quadratic programming (QP) for the safe control gradient $\dot{U}_{\mathrm{safe}}(t_i)$:
\begin{equation} 
\begin{aligned}
    \dot{U}_{\mathrm{safe}}(t_i)
    &= \arg\min_{\dot{U}}
        \big\|\,\dot{U} - \dot{U}_{\mathrm{nominal}}(t_i)\,\big\|, \\
    \text{s.t.}\quad
    &\dot{\phi}(y_i,t_i)
    + \alpha \phi(y_i,t_i)
    + C \phi_0 \le 0,
\end{aligned}
\label{eq:QP}
\end{equation}
where \(\phi_0=\phi(0,Y_0)\) and $\dot{\phi}$ is composed by
\begin{equation}
\dot{\phi} = {\partial}_y \phi \cdot \dot{Y} + \partial_t \phi.
\label{eq:phi_dot}
\end{equation}
Here, $\dot{Y}$ can be obtained from the trained neural operator with the linear dependence on $\dot{U}(t)$, which has been proven by \cite{hu2024safe}.


Given that the BCBF constraint~\eqref{eq:QP} is linear with respect to $\dot{U}$, it can thus be framed and solved by a QP solver.
Given the solved $\dot{U}_{\mathrm{safe}}(t_i)$, we address the potential model mismatch between the trained neural operator $\mathcal{F}_\theta$ and real underlying PDE dynamics with a filtering threshold $\beta$ such that
\begin{equation}
    \dot{U}(t_i) =
    \begin{cases}
    \dot{U}_{\mathrm{safe}}(t_i), 
    & \|\dot{U}_{\mathrm{safe}}(t_i) - \dot{U}_{\mathrm{nominal}}(t_i)\| \le \beta, \\
    \dot{U}_{\mathrm{nominal}}(t_i), 
    & \text{otherwise}.
    \end{cases}
\end{equation}
with the final control signal at the time step $t_i$ is calculated by $u_i= u_{i-1} + \Delta t \cdot \dot{U}(t_i)$.

The online filtering procedures are summarized in \Cref{alg:online-filtering}.
\begin{algorithm}
\caption{Safe Online Filtering with Horizon Agnostic Neural Operator}
\label{alg:online-filtering}
\begin{algorithmic}[1]

    \Require Horizon Agnostic Neural operator $\mathcal{F}:\mathcal{U}\mapsto\mathcal{Y}$, 
             time--dependent CBF $\phi(y,t)$, 
             total time steps $M$, 
             step time interval $\Delta t$
    \Ensure Filtered boundary control sequence $\{\hat{u}_i\}_{i=1}^M$

    \Statex

    \State \textbf{Initialize:}
    \State $\pi(y_i)$ \Comment{The base policy}
    \State $y_i \in \mathbb{R}^n$ \Comment{boundary state at time step $i$}
    \State $u_i \in \mathbb{R}^m$ \Comment{boundary control at time step $i$}
    \State $Y_i = \{y^0, y^1, \dots, y_i\}$ \Comment{state history of one trajectory}
    \State $\hat{Y}_i = \mathcal{F}(U_i)$ \Comment{predicted state from neural operator}

    \Statex

    \For{$i = 1$ \textbf{to} $M$}
        \State $u_i = \pi(y_i)$
        \State $\tilde{U}_i \gets \{\hat{u}_0,\hat{u}_1, \dots,u_i\}$
        \State $\hat{Y}_i \gets \mathcal{F}(\tilde{U}_i)$

        \Statex

        \State Compute CBF time derivative:
        \[
            \dot{\phi}(y_i,t_i)
            = \nabla_y \phi(y_i,t_i)\cdot \dot{Y}_i
              + \partial_t \phi(y_i,t_i)
        \]

        \Statex

        \State Compute safe control rate:
        \[
        \begin{aligned}
            \dot{U}_{\mathrm{safe}}(t_i)
            &= \arg\min_{\dot{U}}
                \big\|\,\dot{U} - \dot{U}_{\mathrm{nominal}}(t_i)\,\big\| \\
            \text{s.t.}\quad
            &\dot{\phi}(y_i,t_i)
            + \alpha \phi(y_i,t_i)
            + C \phi_0 \le 0
        \end{aligned}
        \]

        \Statex

        \If{$\|\dot{U}_{\mathrm{safe}}(t_i) - \dot{U}_{\mathrm{nominal}}(t_i)\|_2 < \beta$}
            \State $\hat{u}_i \gets u_{i-1} + \dot{U}_{\mathrm{safe}}(t_i)\Delta t$
        \Else
            \State $\hat{u}_i \gets u_i$
        \EndIf
        
    \EndFor

\end{algorithmic}
\end{algorithm}

\section{Experimental Results}

\begin{figure*}[h]
    \centering
    \includegraphics[width=\textwidth]{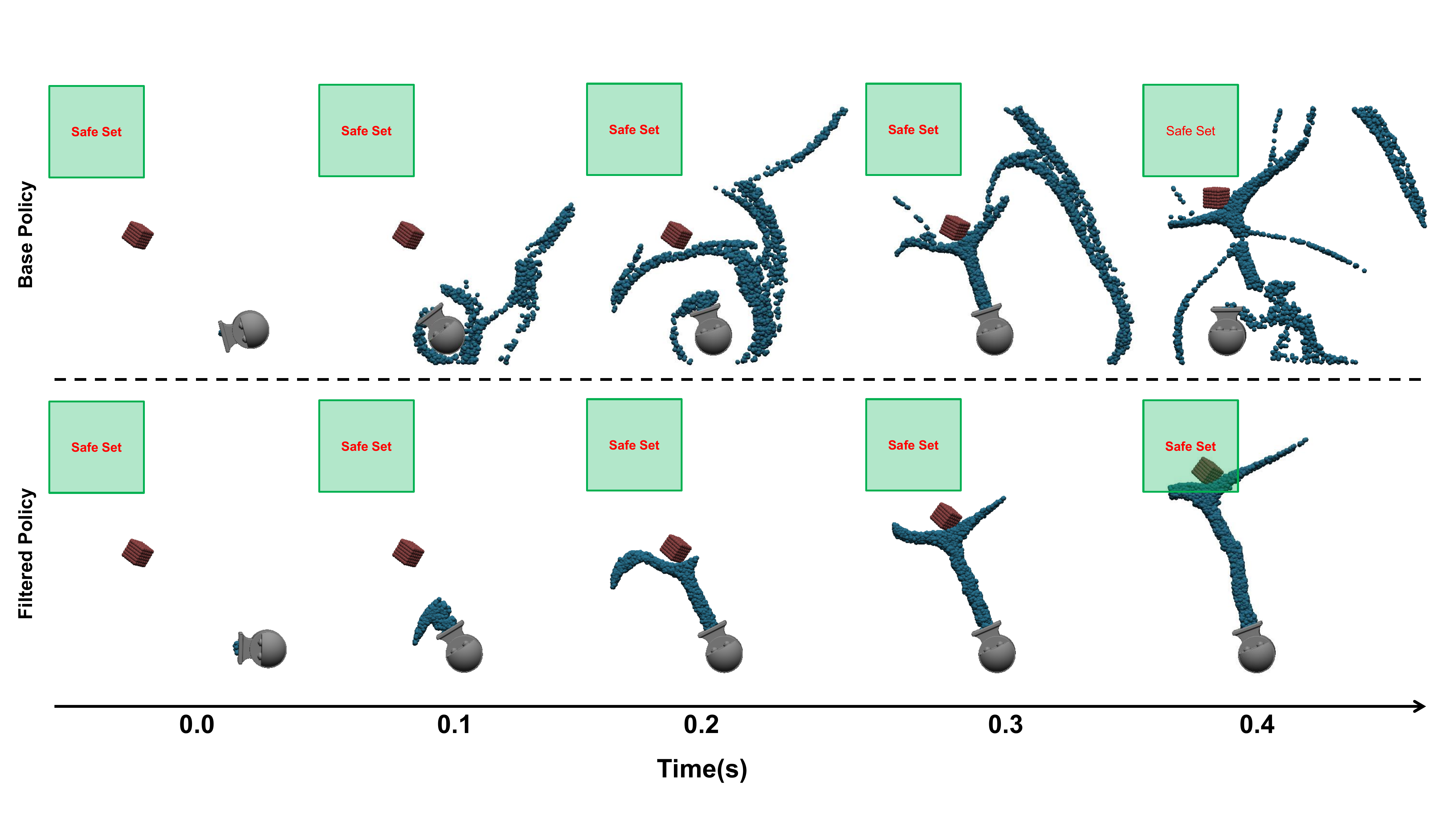}
    \caption{Base and filtered target trajectories in the \emph{transporting} task.}
    \label{fig:transport_xy}
\end{figure*}


\subsection{Experimental Setup}

We evaluate the proposed method in FluidLab~\cite{xian2023fluidlab}, a differentiable
fluid-rigid interaction simulator. 
FluidLab is a testbed for safe boundary
control of PDE-governed robotic manipulation, where coupled
high-dimensional deformable dynamics can be simulated. With this testbed, we learn the input-output operator \(U\mapsto Y\) and train BCBFs from data, and evaluate online filtering under matched nominal/filtered conditions.
In all experiments, we use \(200\) control steps for data collection rollouts and execute real-time control at a frequency of $500$Hz. Trajectory-level and step-level safety metrics are used for analysis.

\subsubsection{Transporting Task}
The \emph{transporting} environment contains a rigid cube interacting with surrounding fluid in a planar boundary-locked workspace. 
The robot agent ejects water with constant speed and is required to transport the target cube into the specified region, and through which the policy is able to control the robot's linear speed on the $x$ axis and the angular speed, which we denote by \(U(t)\in\mathbb{R}^2\). The task-relevant boundary state
\(Y(t)\) is constructed from cube planar coordinates and used by the learned model and BCBF. The safe set is defined as a geometric
target region in the \(x\)-\(y\) plane:
\[
\mathcal{S}_0^{\mathrm{trans}}
=
\{Y: x\in[0.05,0.25],\; y\in[0.65,0.95]\}.
\]
A rollout is safe at a step if the cube center lies in \(\mathcal{S}_0^{\mathrm{trans}}\);
distance to this set is additionally recorded as a continuous safety margin.

\subsubsection{Gathering Task}
The gathering environment contains two rigid bodies immersed in fluid, and the
robot equipped with a rigid spatula that is able to paddle the water and specified to reduce the rigid bodies' separation. The policy outputs boundary control signal \(U(t)\in\mathbb{R}^2\) which commands its linear speed on $X, Y$ plane. The boundary state \(Y(t)\) is defined from object-relative
quantities (difference of object positions) and, in the scalar setting, by an
auxiliary safety variable \(s(t)\) that is the distance between the two bodies. The safe set is
defined by a distance threshold:
\[
\mathcal{S}_0^{\mathrm{gather}}
=
\{Y: \|p_1-p_2\|_2 < \delta\}, \quad \delta=0.2.
\]

\subsection{Implementation Details}

\textbf{Base policy and data generation.}
The base policy is a noisy proportional controller with task-specific geometric priors.
Let \(\mathrm{clip}_{[\ell,u]}(\cdot)\) denote component-wise clipping to
\([\ell,u]\), and let \(\eta_i\sim\mathcal{N}(0,\Sigma)\) denote injected control
noise.

For \emph{transporting}, define robot planar position \(p_i^r=[x_i^r,y_i^r]^\top\),
yaw \(\psi_i\), and target point \(p_i^c=[x_i^c,y_i^c]^\top\) (cube-related reference
from the simulator state). The two effective control channels are forward velocity
and yaw rate, \(U_i=[v_i,\omega_i]^\top\). The translational component is
\begin{equation}
v_i=
\begin{cases}
\mathrm{clip}_{[-v_{\max},v_{\max}]}\!\left(k_p(x_i^c-x_i^r)+\eta_{x,i}\right),
& x_i^r < x_i^c,\\[2mm]
\mathrm{clip}_{[-v_{\max},v_{\max}]}\!\left(k_p(x_i^c-x_i^r)+\eta_{x,i}\right),
& x_i^r > x_i^c+ d,\\[2mm]
0, & \text{otherwise},
\end{cases}
\label{eq:exp_transport_v}
\end{equation}
where \(d>0\) is the distance margin that keeps the robot
behind the cube before pushing and $k_p$ is the gain. The yaw channel aligns the robot heading to the target:
\begin{equation}
\psi_i^\star=\operatorname{atan2}(y_i^c-y_i^r,\;x_i^c-x_i^r),\qquad
e_{\psi,i}=\operatorname{mod}(\theta+\pi,2\pi)-\pi,
\end{equation}
\begin{equation}
\omega_i=
\begin{cases}
\mathrm{clip}_{[-\omega_{\max},\omega_{\max}]}\!\left(k_\omega e_{\psi,i}+\eta_{\omega,i}\right),
& |e_{\psi,i}|>\varepsilon_\psi,\\
0, & \text{otherwise}.
\end{cases}
\label{eq:exp_transport_w}
\end{equation}
where $e_{\psi,i}$ is the mismatch between robot's heading and the actual robot-to-target angle. In our implementation, \(v_{\max}=\omega_{\max}=1\), \(d=0.15\),
and \(\varepsilon_\psi=0.05\).

For \emph{gathering}, the robot tracks an elliptical reference trajectory in the
horizontal plane. Let \((x_c,y_c)\) be the ellipse center, \(a,b\) the semi-axes,
\(\vartheta\) the ellipse orientation, and \(\rho\) the speed-scale factor.
With \(\theta_i\in[0,2\pi\rho]\), the reference is
\begin{equation}
\begin{bmatrix}
x_i^{\mathrm{ref}}\\
y_i^{\mathrm{ref}}
\end{bmatrix}
=
\begin{bmatrix}
x_c\\y_c
\end{bmatrix}
+
\begin{bmatrix}
\cos\vartheta & -\sin\vartheta\\
\sin\vartheta & \cos\vartheta
\end{bmatrix}
\begin{bmatrix}
a\cos\theta_i\\
b\sin\theta_i
\end{bmatrix}.
\label{eq:exp_gather_ref}
\end{equation}
The effective boundary control \(U_i=[v_{x,i},v_{y,i}]^\top\) is
\begin{equation}
\begin{aligned}
v_{x,i} &= \mathrm{clip}_{[u_{\min},u_{\max}]}\!\left(
k_p(x_i^{\mathrm{ref}}-x_i^r)+\eta_{x,i}
\right), \\
v_{y,i} &= \mathrm{clip}_{[u_{\min},u_{\max}]}\!\left(
k_p(y_i^{\mathrm{ref}}-y_i^r)+\eta_{y,i}
\right).
\end{aligned}
\label{eq:exp_gather_pd}
\end{equation}

\begin{figure*}[h]
    \centering
    \includegraphics[width=\textwidth]{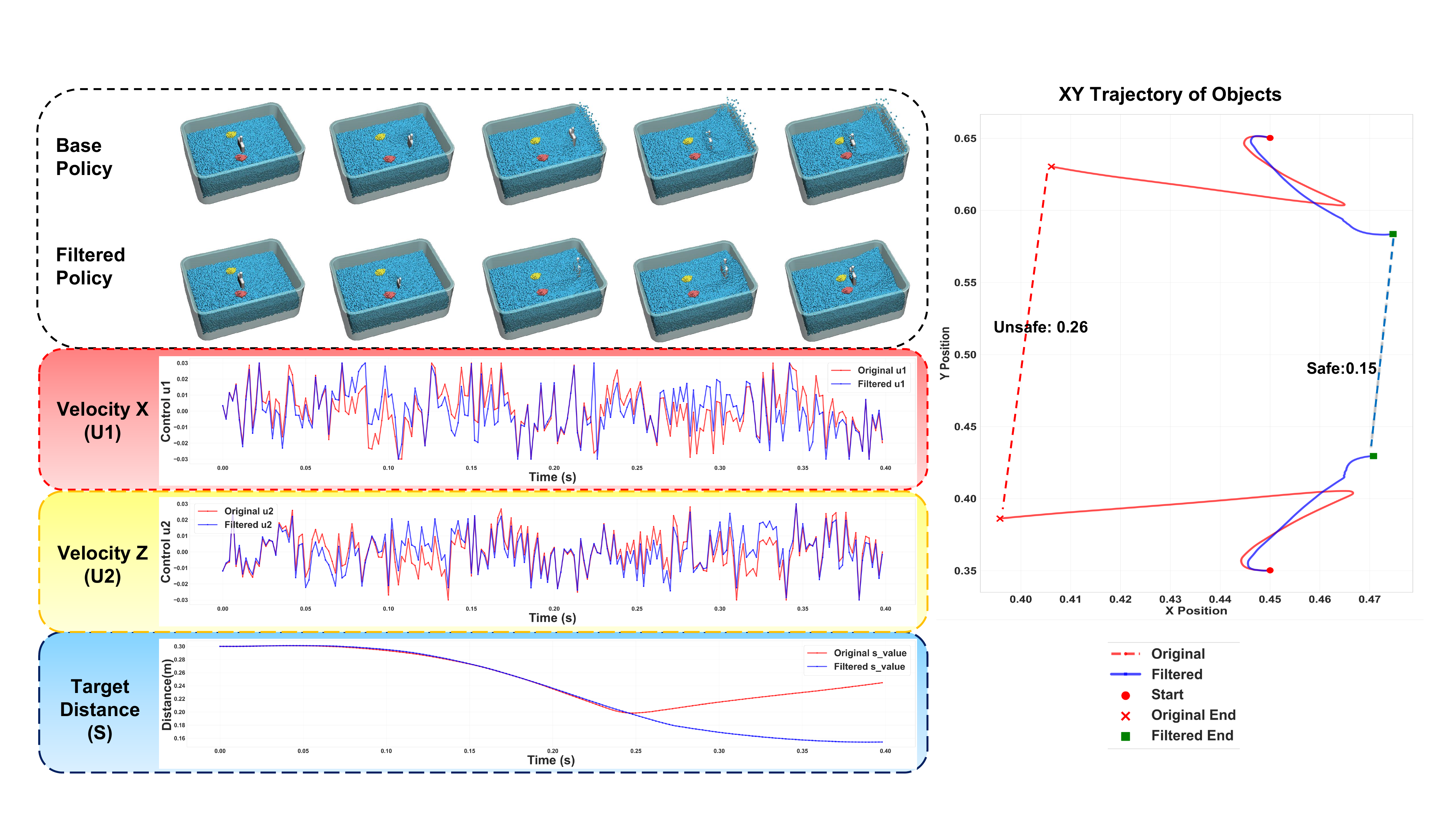}
    \caption{Comparison between the base policy and the filtered safe policy on the \emph{gathering} task. The original and filtered control inputs, the evolution of the target distance, and the object trajectories in the XY plane are presented. The base policy induces turbulence near the end of the task, which pushes the objects farther apart and increases the distance around $t \approx 0.25$s. The filtered policy mitigates this disturbance, maintains convergence toward the desired distance, and improves overall safety performance.} 
    \label{fig:gathering_result}
\end{figure*}
\textbf{Horizon-agnostic neural operator training.}
The NO training pipeline uses preprocessed task trajectories and selects
\(n_{\mathrm{train}}\) and \(n_{\mathrm{test}}\) episodes for training and testing,
respectively. Each episode has horizon
\(M=200\) before optional temporal subsampling. Let one batch contain episodes
\(\{(U_b,Y_b)\}_{b=1}^{B}\), where \(U_b\in\mathbb{R}^{M\times 2}\) and
\(Y_b\in\mathbb{R}^{M\times d_y}\) (task-dependent \(d_y\)). The collate function
samples one prefix length per batch
\begin{align}
\tilde{M} &\sim \mathcal{U}\{M_{\min},\ldots,M_{\max}\},
\end{align}
and crops all trajectories to the first \(\tilde{M}\) steps
\[
\tilde{U}_b=U_b[1\!:\!\tilde{M}],\qquad
\tilde{Y}_b=Y_b[1\!:\!\tilde{M}].
\]
Hence, all samples in one batch share the same length, while \(\tilde{M}\) changes
across batches, which directly realizes horizon-agnostic training.
The Fourier neural operator used has $64$ hidden channels, $4$ layers and the frequency latent embeddings is truncated to keep the lowest $16$ frequency modes. The model is trained using the AdamW optimizer with a learning rate of $1\mathrm{e}{-}3$ and a weight decay of $1\mathrm{e}{-}3$.

\textbf{BCBF network and optimization.}
The BCBF model is an MLP
\(\phi_\psi:\mathbb{R}^{1+d_y}\rightarrow\mathbb{R}\) with hidden widths
\([32,128,64,32]\), where the input is \([t,Y]\). Training data are the same task
rollouts sampled by
\(n_{\mathrm{train}}\), \(n_{\mathrm{test}}\), reshaped over all time points in each
trajectory. For each batch, we construct
\[
\mathbf{y}_{tt}=[t,Y],\quad
\mathbf{U}_0=[t_0,Y_0],\quad
\dot{\mathbf{Y}}=\left[1,\frac{dY}{dt}\right],
\]
where \(dY/dt\) is computed through finite difference.
We followed the BCBF loss formulation in \cite{hu2024safe}, with  
NAdam optimizer with betas \(0.9,0.999\), weight decay \(0.1\), and learning rate \(0.01\).

\subsection{Results}

\begin{table}[t]
\centering
\caption{Safety and convergence comparison between nominal and filtered policies.}
\label{tab:main_results}
\begin{tabular}{|l|c|c|}
\hline
\textbf{Metric} & \textbf{Transporting} & \textbf{Gathering} \\
\hline
Base policy safe rate & 57.8\% & 45.5\% \\
Offline filter safe rate & 18.2\% (\(\downarrow 39.6\)) & 27.3\% (\(\downarrow 18.2\)) \\
Filtered policy safe rate & \textbf{80.2\%} (\(\uparrow 22.3\)) & \textbf{63.6\%} (\(\uparrow 18.1\)) \\
\hline
Base avg. unsafe steps & 179.18 & 166.81 \\
Offline filter avg. unsafe steps & 195.51 (\(\uparrow 16.33\)) & 186.30 (\(\uparrow 19.49\)) \\
Filtered avg. unsafe steps & \textbf{169.54} (\(\downarrow 9.64\)) & \textbf{160.36} (\(\downarrow 6.45\)) \\
\hline
\end{tabular}
\end{table}


Table~\ref{tab:main_results} shows safety and convergence comparison between nominal and filtered policies, where online filtering consistently improves safety in both tasks at the trajectory level. In \emph{transporting}, the filtered policy increases the safe-trajectory rate from \(57.8\%\) to \(80.2\%\), which is a gain of \(22.4\%\) over the base policy. \Cref{fig:transport_xy} shows the base and filtered target trajectories, where the base policy results in stagnation and the filtered policy successfully transported to the set of targets. 
In \emph{gathering}, the safe-trajectory rate increases from \(45.5\%\) to \(63.6\%\), which is an improvement of \(18.1\%\). These gains indicate that the filter improves the probability that an entire rollout remains or becomes safe under the same base policy backbone. 
\Cref{fig:gathering_result} visualizes the control inputs, distance evolution, and the trajectories of the two target objects in the XY plane. The base policy fails to converge to the required target distance and, at a later stage, pushes the objects farther apart. This behavior is confirmed by the distance evolution plot in \Cref{fig:gathering_result}, where the distance increases around $t \approx 0.25$s. 
To better understand this phenomenon, we examine simulation snapshots sampled at 0.1s intervals. Around $t = 0.3$s, the base policy generates noticeable turbulence while maneuvering the spatula back toward the paddle position. This disturbance induces additional water flow that pushes the objects away from each other, preventing convergence. In contrast, the filtered policy avoids this behavior by adaptively steering the spatula to mitigate water disturbance, thereby maintaining stable convergence toward the desired distance.

At the step level, the filtered policy also reduces the average number of steps required to reach convergence to the safe condition. In \emph{transporting}, this metric drops from \(179.18\) to \(169.54\) (\(9.64\) fewer steps, \(\approx 5.4\%\) reduction). In \emph{gathering}, it decreases from \(166.81\) to \(160.36\) (\(6.45\) fewer steps, \(\approx 3.9\%\) reduction). Since trajectories that reach the safe set earlier spend fewer timesteps in unsafe transients, this reduction provides evidence of improved cumulative step-level safety behavior in addition to higher trajectory-level safety rates. 

Moreover, for both \emph{transporting} and \emph{gathering}, we observe that applying a neural operator trained offline as a fixed filter degrades performance, as shown in Table~\ref{tab:main_results}. The safety rate decreases, and convergence requires more steps than the base policy. This result highlights that the safety filter must be adapted online to the current base policy and environment dynamics in order to provide consistent improvement.


Taken together, these results demonstrate the effectiveness of the proposed online filter, across different safety geometries including region-reaching safety (\emph{transporting}) and pairwise-distance safety (\emph{gathering}). The proposed method improves rollout-level safety success rates while simultaneously reducing cumulative unsafe exposure over time.

\section{Conclusion}

This paper presents an online safety filtering framework for manipulating deformable objects whose dynamics is governed by a PDE system. We combine a horizon-agnostic neural operator for variable-horizon boundary input-output modeling with a BCBF-based
quadratic-program safety filter that runs at each control cycle. The resulting
pipeline preserves the nominal policy structure while enforcing task-level safety
constraints through minimal control modification. Experiments on FluidLab
transporting and gathering tasks show improved safe-trajectory rates and reduced
steps to safety convergence compared with the nominal policy.
Future work will study transfer to broader deformable manipulation domains and hardware-in-the-loop deployment with tighter real-time guarantees.


\bibliographystyle{IEEEtran}
\bibliography{references}

\newpage

\vfill

\end{document}